\documentclass{article}
\usepackage{spconf}
\usepackage{times}
\usepackage{latexsym}
\usepackage{amsmath}
\usepackage{amssymb}
\usepackage{graphicx}
\usepackage{makecell}
\usepackage[table]{xcolor}
\usepackage{colortbl}
\usepackage[hidelinks]{hyperref}

\title{Continual Learning in Machine Speech Chain \linebreak Using Gradient Episodic Memory}
\name{
    \begin{tabular}{c}
    Geoffrey Tyndall\textsuperscript{\normalfont 1}\sthanks{This work was conducted while the first author was doing internship at HA3CI Laboratory, JAIST, Japan under JST Sakura Science Program}, Kurniawati Azizah\textsuperscript{\normalfont 1}, Dipta Tanaya\textsuperscript{\normalfont 1},
    \\
    Ayu Purwarianti\textsuperscript{\normalfont 2}, Dessi Puji Lestari\textsuperscript{\normalfont 2}, Sakriani Sakti\textsuperscript{\normalfont 3,4}
    \end{tabular}
}

\address{
    \textsuperscript{1}University of Indonesia, Indonesia
    \\
    \textsuperscript{2}Bandung Institute of Technology, Indonesia
    \\
    \textsuperscript{3}Nara Institute of Science and Technology, Japan
    \\
    \textsuperscript{4}Japan Advanced Institute of Science and Technology, Japan
    \\
    \normalsize \texttt{geoffrey.tyndall91@ui.ac.id,}
    \\
    \normalsize \texttt{\{kurniawati.azizah,diptatanaya\}@cs.ui.ac.id,}
    \\
    \normalsize \texttt{\{ayu,dessipuji\}@itb.ac.id, ssakti@is.naist.jp}
}
\begin{document}
\maketitle
\begin{abstract}
Continual learning for automatic speech recognition (ASR) systems poses a challenge, especially with the need to avoid catastrophic forgetting while maintaining performance on previously learned tasks. This paper introduces a novel approach leveraging the machine speech chain framework to enable continual learning in ASR using gradient episodic memory (GEM). By incorporating a text-to-speech (TTS) component within the machine speech chain, we support the replay mechanism essential for GEM, allowing the ASR model to learn new tasks sequentially without significant performance degradation on earlier tasks. Our experiments, conducted on the LJ Speech dataset, demonstrate that our method outperforms traditional fine-tuning and multitask learning approaches, achieving a substantial error rate reduction while maintaining high performance across varying noise conditions. We showed the potential of our semi-supervised machine speech chain approach for effective and efficient continual learning in speech recognition. 
\end{abstract}
\section{Introduction}
\label{sec:intro}
The exceptional performance of deep learning architectures, as illustrated by the Transformer model \cite{vaswani:transformer}, has enabled state-of-the-art automatic speech recognition (ASR) systems to reach levels of accuracy comparable to human performance \cite{li:jasper-cnn-asr, zhang:transformer-asr, gulati:conformer}. These advancements have significantly enhanced speech recognition capabilities. However, a critical challenge persists: ASR systems should be capable of recognizing a continuous stream of tasks. Despite the existence of large-scale speech models \cite{radford:whisper, seamless:seamlessm4t} that excel in multitask performance, these models demand substantial resources in terms of data and computational power, and they require the availability of all tasks from the beginning, i.e., offline learning.

An alternative approach to this issue is \textit{fine-tuning}, which transfers knowledge from one task to another, or \textit{multitask learning}, where the model is trained from scratch using both previous and new task data simultaneously. Unfortunately, the former approach (\textit{transfer learning}) can degrade the model's performance on earlier tasks due to \textit{catastrophic forgetting} \cite{mccloskey:catastrophic-forgetting}. Meanwhile, the latter approach necessitates retaining old data to mix with new task data, potentially leading to privacy concerns.

Continual learning is a paradigm designed to allow models to learn new tasks sequentially without compromising their ability to perform previous tasks or violating data privacy. Its effectiveness in sequentially handling multiple recognition tasks was recently demonstrated in \cite{chang:lifelong-learning-asr}.

Unlike existing fully supervised methods for conducting continual learning experiments on ASR, this paper proposes a semi-supervised approach within the machine speech chain framework \cite{tjandra:msc-journal-published}. Our method integrates text-to-speech (TTS) to support a replay mechanism in continual learning. We adopt gradient episodic memory (GEM) \cite{lopez-paz:gem} as our chosen implementation for this replay-based continual learning scenario.

We evaluate our proposed method against other prevalent learning paradigms such as fine-tuning and multitask learning. Our results indicate that continual learning within the machine speech chain framework offers superior performance compared to these traditional methods and serves as a viable alternative to fully supervised continual learning. Although the upper bound fully supervised continual learning achieves a lower error rate, our approach manages to achieve a 40\% average error rate reduction relative to fine-tuning. Therefore, our contributions include: (1) proposing a machine speech chain-based method for enabling continual learning in speech recognition; (2) conducting experiments to validate our method using the LJ speech dataset.

\begin{figure}
    \centering
    \includegraphics[width=1\linewidth]{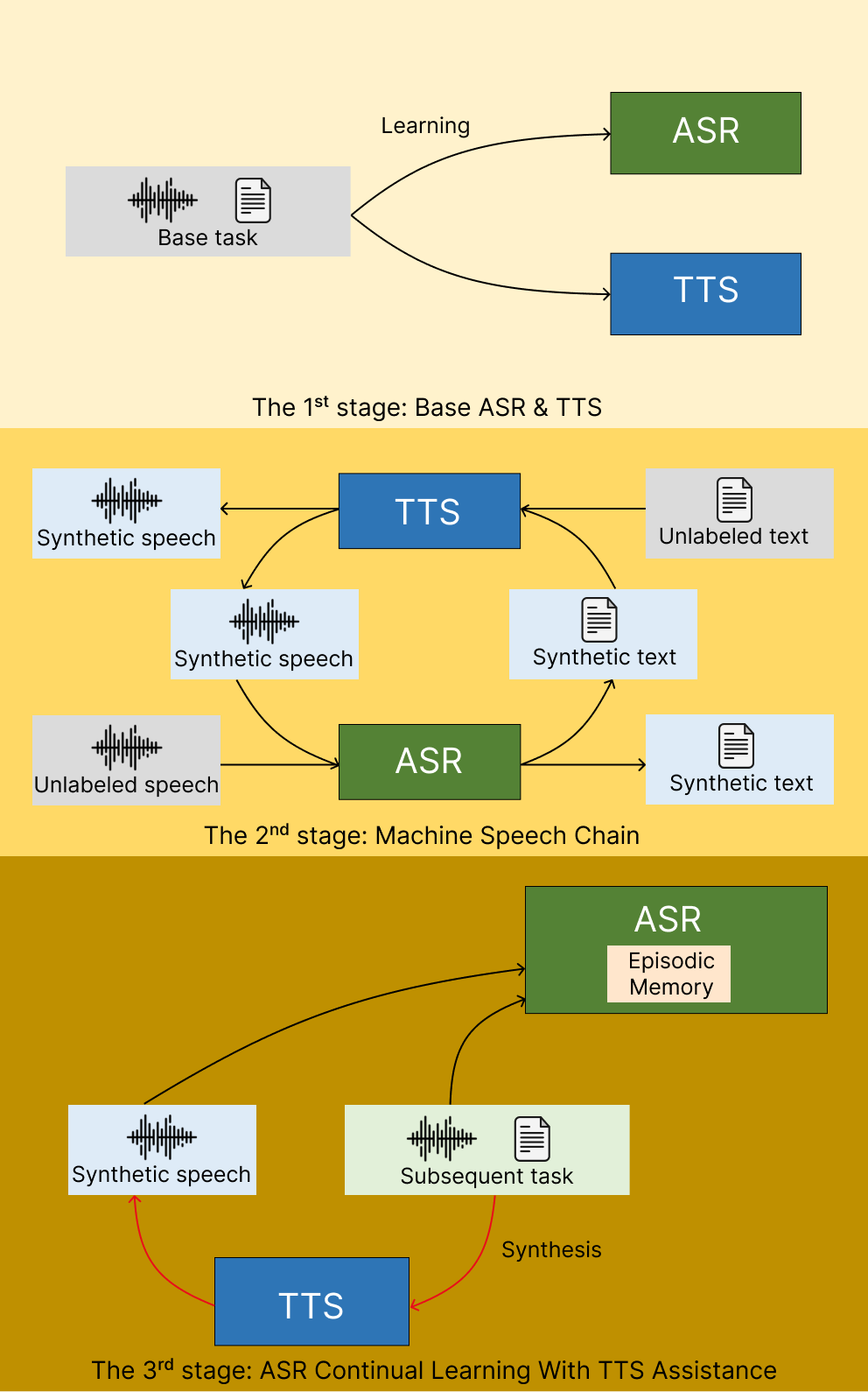}
    \caption{Continual learning in the machine speech chain framework.}
    \label{fig:stages}
\end{figure}
\section{Related Work}
\label{sec:related-work}

\subsection{Machine Speech Chain}
\label{sec:machine-speech-chain}
Machine speech chain is an architecture that connects sequence-to-sequence model of automatic speech recognition (ASR) and text-to-speech (TTS) in a closed-loop framework. This integration was proposed to be representative of human speech chain mechanism \cite{tjandra:msc-journal-published}, which is listening while speaking \cite{denes:speech-chain}. To date, machine speech chain has been used in various works, adaptive lombard TTS \cite{novitasari:tts-dynamic-adaptive}, data augmentation \cite{qi:SpeeChain-toolkit}, and code-switching \cite{vaza:code-switching-msc}.

\subsection{Gradient Episodic Memory}
\label{sec:gradient-episodic}
Gradient episodic memory (GEM) is a replay method of continual learning paradigm \cite{lopez-paz:gem}. GEM exploits samples from the past task's data when encountering the data of a new task to minimize the L2 distance between the gradients of the new task's data and the old ones' data, i.e.,
\begin{equation}\label{eq:gem-projection}
    \begin{split}
    & \min_{\boldsymbol{\Tilde{g}}} \frac{1}{2}\|\boldsymbol{g - \Tilde{g}}\|^{2}_{2}
    \\
    & \text{s.t.} \langle\boldsymbol{\Tilde{g}},\boldsymbol{g_{k}}\rangle \geq 0, \forall k \in (0,...,i-1),
    \end{split}
\end{equation}
where $\boldsymbol{g},\boldsymbol{\Tilde{g}_{k}} \in \mathbb{R^{|\theta|}}$ and $|\theta|$ is the number of model's parameters. ASR model that was equipped with GEM in previous finding (see \cite{chang:lifelong-learning-asr}) outperformed regularization-based methods , such as synaptic intelligence \cite{zenke:synaptic}, or knowledge distillation \cite{hinton:knowledge-distillation}, in a continual learning scenario with different acoustic and topic domain acted as task boundary. In this paper, we introduce GEM usage in machine speech chain and we demonstrate first-hand to show its potential.

\section{Machine Speech Chain Using GEM}
\label{sec:machine-speech-chain-using-gem}
We introduce a three-stage mechanism designed to enable ASR models to perform continual learning in a semi-supervised manner, achieving satisfactory results with minimal forgetting. These three stages, depicted in Figure \ref{fig:stages}, build upon the process proposed in \cite{tjandra:msc-journal-published}, for the first and second stages, with our continual learning method introduced in the third stage.

\begin{enumerate}
    \item \textbf{First stage}: Supervised learning on the base task. Here, ASR and TTS are trained separately in a supervised manner to ensure strong baseline performance for the subsequent training stages.
    \item \textbf{Second stage}: Semi-supervised learning. At this stage, ASR and TTS mutually enhance each other by training on unlabeled data from the base task, using unsupervised methods to improve performance.
    \item \textbf{Third stage}: Continual learning. ASR engages in continual learning for new tasks using replayed inputs from the base task, synthesized by TTS.
\end{enumerate}

In our approach, the replay process for speech recognition leverages TTS as a synthesis model to generate pseudo-samples of the base task. These pseudo-samples are stored in episodic memory and used by GEM to regulate the gradients for both new and previous tasks.

During the third stage, when the machine speech chain encounters incoming tasks as:
\begin{equation}\nonumber
    \begin{split}
    [D^{1},...,D^{n}] & = [(x^{1},y^{1}),...,(x^{n},y^{n})] 
    \\
    & = [\{(x_{1}^{1},y_{1}^{1},...,(x_{|D^{1}|}^{1},y_{|D^{1}|}^{1})\},...,
    \\
    & \hspace{0.6cm} \{(x_{1}^{n},y_{1}^{n}),...,(x_{|D^{n}|}^{n},y_{|D^{n}|}^{n})\}],
    \end{split}
\end{equation}
where $x$ is the input and $y$ is the label, we forward the speech data label to TTS to generate pseudo-samples of the base task, i.e., $\hat{x}^{0} \sim TTS(y^i)$. These synthesized samples are stored, along with the data from the incoming task, processed as follows:
\begin{align}
        \mathcal{M}_{0} \leftarrow \mathcal{M}_{0} \cup (\hat{x}^{0},y^{i})
        \\
        \mathcal{M}_{i} \leftarrow \mathcal{M}_{i} \cup (x^{i},y^{i})
        \\
        g \leftarrow \nabla_{\theta}\ell(ASR_{\theta}(x^{i}), y^{i})
        \\
        g_{k} \leftarrow \nabla_{\theta}\ell(ASR_{\theta},\mathcal{M}_{k})  \text{ for all } k < i
        \\
        \Tilde{g} \leftarrow \text{PROJECT}(g,g_{0},g_{1},...,g_{i-1})\text{, see (\ref{eq:gem-projection})}
        \\
        \theta \leftarrow \theta - \delta \Tilde{g},
\end{align}
where $\mathcal{M}_{}$ represents the episodic memory and $\delta$ denotes the weight assigned for updating the model parameters during continual learning for the $i$-th task ($i > 0$).

To our knowledge, our proposed mechanism is the first to incorporate TTS within the continual learning framework for ASR. While prior works in continual learning have utilized various generative models \cite{shin:continual-generative-replay, atkinson:pseudo-recursal}, none has specifically employed TTS for continual learning in ASR.

\begin{table}
\centering
\begin{tabular}{lrr}
\hline
\textbf{Model} & \thead{\textbf{CER (\%)} \\ \textbf{LJ Original}} & \thead{\textbf{CER (\%)}\\ \textbf{LJ Noisy} }\\
\hline
\textbf{ASR$_{\text{Lower}}$}\\
\cellcolor[HTML]{FFE6CC} Pre-trained & 9.2 &  82.6 \\
Fine-tuning & 19.0 & 31.3 \\
GEM & 8.5 & 15.8 \\
Multitask & 74.8 & 76.7 \\
\hline
\textbf{ASR$_{\text{SpeechChain}}$}\\
\cellcolor[HTML]{FFD966} Pre-trained & 6.4 &  95.7 \\
Fine-tuning & 12.7 & 33.1 \\
\cellcolor[HTML]{BF9000} GEM & 11.1 & 15.5 \\
\hline
\textbf{ASR$_\text{Upper}$}\\
Pre-trained & 1.9 &  108.4 \\
Fine-tuning & 6.7 & 15.6 \\
GEM & 5.2 & 8.4 \\
Multitask & 3.8 & 10.9 \\
\hline
\end{tabular}
\caption{CER results for different methods applied on the ASR model. The color-coded rows (\textcolor[HTML]{FFE6CC}{$\blacksquare$} $1^{\text{st}}$, \textcolor[HTML]{FFD966}{$\blacksquare$} $2^{\text{nd}}$, \textcolor[HTML]{BF9000}{$\blacksquare$} $3^{\text{rd}}$) represent each stage of our proposed machine speech chain-based method.}
\label{tab:result}
\end{table}

\section{Experiments}
\label{sec:experiments}
\begin{table}
    \centering
    \begin{tabular}{c r r}
    \hline
    \thead{\textbf{Split Ratio} \\ \textbf{Labeled / Unlabeled}} & \thead{\textbf{CER (\%)} \\ \textbf{LJ Original}} & \thead{\textbf{CER (\%)}\\ \textbf{LJ Noisy}}\\
    \hline
    30 / 70 & 11.1 & 15.5 \\
    50 / 50 & 4.8 & 11.5 \\
    70 / 30 & 4.0 & 10.9 \\
    \hline
    \end{tabular}
    \caption{Results for the \textbf{ASR${_\text{SpeechChain}}$} with updates of different ratio of labeled and unlabeled data during base task learning in the first stage and second stage of the framework.}
    \label{tab:result-split}
\end{table}
\begin{figure*}
    \centering
    \includegraphics[width=1\linewidth]{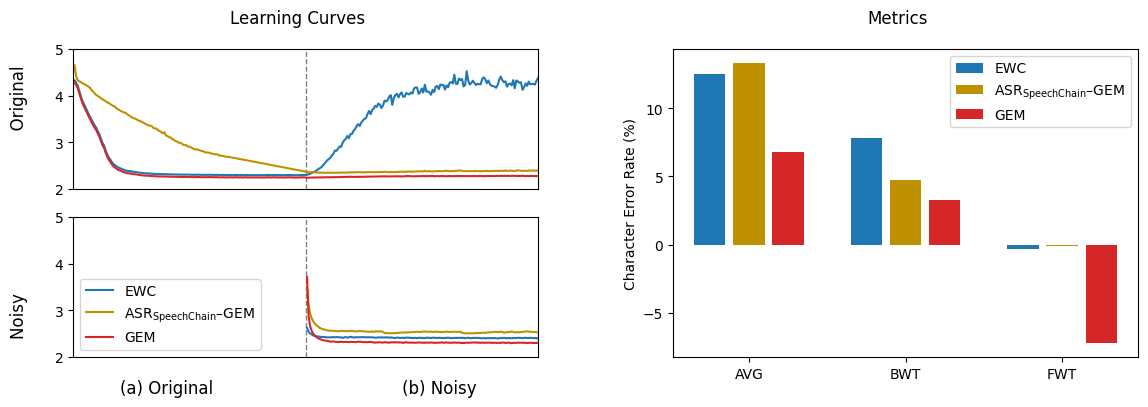}
    \caption{Learning curves of models in continual learning paradigm and their respective metrics.}
    \label{fig:continual-comparison}
\end{figure*}

\subsection{Experimental Setup}
\label{sec:experimental-setup}
We prepared two tasks for the ASR models to recognize. The first task, referred to as the base task, utilized the clean original dataset of LJ Speech \cite{ito:ljspeech}, consisting of 24 hours of audio. To simulate different scenario for the subsequent task, we created a noisy version of the original speech dataset. This noisy dataset also comprises 24 hours of audio, but with added white noise at a signal-to-noise ratio (SNR) of 0. Consequently, the base task is denoted as LJ Original, and the subsequent task is denoted as LJ Noisy. Both datasets were split into train, dev, and test sets with a ratio of 94\%, 3\%, and 3\%, respectively.

For the ASR architecture, we employed the Speech-Transformer \cite{dong:speech-transformer}, while the TTS architecture was based on the Transformer-based Tacotron 2 \cite{li:tacotron2-transformer}. All of the ASR models did not involve hyperparameter tuning since they already employed almost identical hyperparameters to those that had been used in \cite{dong:speech-transformer}. The architecture of the ASR models employed 12 encoder blocks, 6 decoder blocks, 4 attention heads, and a feed-forward hidden layer size of 2048. We used 80 dimensions for the Mel-spectrogram input. We trained the models using the Adam optimizer with $\beta_{1} = 0.9$, $\beta_{2} = 0.98, \epsilon = 10^{-9}$ and employed cross-entropy loss with neighborhood smoothing. The episodic memory that we used for continual learning had size of 100 samples per task, or in other word 1\% of dataset size. 

For TTS models that are needed in machine speech chain condition, we configured them to be consisted of 6 encoder blocks for the transformer-based encoder, 6 decoder blocks for the autoregressive decoder, 8 heads, and a feed-forward hidden layer size of 2048. These values were identical to the best configuration that had been used in \cite{li:tacotron2-transformer}. The TTS input was the character sequence, and the output was the 80 dimensions of the Mel-spectrogram. We used the Adam optimizer with the same $\beta_{1}, \beta_{2}, \epsilon$ values and employed cross-entropy loss.

Our experiment involved training ASR models under supervised conditions: lower bound and upper bound, and our proposed method that involved semi-supervised condition: machine speech chain. The upper and lower bound refers to the amount of base task data provided to the ASR model before it engages in learning with subsequent task. Specifically, we varied the proportion of the LJ Original training data while keeping the LJ Noisy training data constant at 100\% of the train set. We used 30\% of LJ Original train set for the lower bound condition, 30\% of the train set as labeled data \& 70\% of the train set as unlabeled data for the machine speech chain condition, and 100\% of the train set for the upper bound condition.

\subsection{Experiment Result}
\label{sec:experiment-result}

\subsubsection{Continual Learning Performance}
\label{sec:continual-learning-performance}

The experimental results, as detailed in Table \ref{tab:result}, demonstrate the efficacy of various continual learning approaches applied to the ASR model in both clean (LJ Original) and noisy (LJ Noisy) conditions. The \textbf{ASR$_{\text{Lower}}$} results show that the GEM approach significantly reduces the character error rate (CER) compared to fine-tuning and multitask learning. For instance, GEM achieved a CER of 8.5\% on LJ Original and 15.8\% on LJ Noisy, outperforming the fine-tuning method which resulted in CERs of 19.0\% and 31.3\% respectively. Multitask learning, however, showed the highest CERs of 74.8\% and 76.7\%, indicating its limitation in handling noise without optimal balance of data.

The \textbf{ASR${_\text{SpeechChain}}$} model trained with GEM outperformed the fine-tuning method, achieving CERs of 11.1\% and 15.5\% for LJ Original and LJ Noisy respectively. This is a significant improvement over fine-tuning, which recorded CERs of 12.7\% and 33.1\%. Furthermore, comparing the GEM method across different models, \textbf{ASR${_\text{Upper}}$} using GEM achieved the lowest CERs at 5.2\% and 8.4\%, compared to fine-tuning and multitask methods. However, it is important to highlight that the ASR${_\text{SpeechChain}}$ model, despite not reaching the lowest error rates, still showed substantial improvements. The ASR${_\text{SpeechChain}}$ model with GEM achieved significant error rate reductions, comparable to the ASR$_{\text{Upper}}$ model, with a 40\% error rate reduction relative to the respective fine-tuning methods. We also demonstrate the results with different split ratio of labeled and unlabeled data of the base task in Table \ref{tab:result-split}, where we can observe that with increasing labeled data the error rates are becoming smaller. These results emphasize that our proposed method is effective, mitigating catastrophic forgetting and maintaining consistent performance across tasks and varying semi-supervised learning scenarios.

\subsubsection{Continual Learning Comparison}
\label{sec:continual-learning-comparison}
We also compared our semi-supervised method to the other continual learning methods which are carried out in a fully supervised scenario. These other methods were gradient episodic memory (GEM) and elastic weight consolidation (EWC) \cite{kirkpatrick:ewc}. We can see from Figure \ref{fig:continual-comparison} that the learning curves exhibit the superiority of GEM, as models that leveraged GEM as their replay process were able to prevent catastrophic forgetting. Although EWC had worse forgetting prevention, it performed better on learning new task because of its fully supervised scenario.

We also computed the continual learning metrics, such as average (AVG), backward transfer (BWT), and forward transfer (FWT) character error rate, as shown in Figure \ref{fig:continual-comparison}, which were useful for comparing the three models to each other. In our experiment, BWT is defined as the ability of a model to transfer the lowest possible error to the previous task it has encountered, while FWT is defined as the ability of a model to learn a new task with the lowest possible error compared to the error rate attained by the standard fine-tune method.

GEM, when applied in a supervised ASR system, as expected, achieved the lowest of all the metrics. EWC had a slightly lower AVG at 12.5\% than ASR${_\text{SpeechChain}}$, which achieved 13.3\%. Our model performed well in reducing forgetting by introducing a lower error to the previous task with a BWT at 4.7\% than EWC's at 7.8\%. For the FWT metric, our model and EWC performed relatively similarly at -0.3\% and -0.1\% respectively. From these results, we can observe that our model works as intended to learn sequential tasks, prevent catastrophic forgetting, and exploit accumulated knowledge to learn a new task, which are all the properties of a functioning continual learning process.

\section{Conclusion}
\label{sec:conclusion}
We proposed a novel method to allow automatic speech recognition (ASR) model to perform continual learning in a semi-supervised manner of machine speech chain. We then demonstrated first-hand the implementation of such replay method with gradient episodic memory (GEM). Although our upper bound supervised model achieved lower CER than our proposed method, the machine speech chain-based method managed to get the same 40\% averaged error rate reduction. Furthermore, we compared both machine speech chain that was trained under the proposed continual learning scenario with the machine speech chain under the fine-tuning scenario. We found that our method worked and achieved minimal forgetting, or prevented catastrophic forgetting. This showed that our novel method has potential for further application of speech recognition and can serve as an alternative to the fully supervised mechanism of continual learning. We believe this paper provides the first exploration of continual learning in machine speech chain framework and makes a step towards realizing effective and efficient learning for speech recognition.

\section{Limitations}
\label{sec:limitations}
We acknowledge the need for further experiments to assess the generalizability of our approach. While this work demonstrates success on a simple task boundary of noise variation, future work will involve applying our method to a wider range of tasks, such as multilingual speech recognition (where the model needs to adapt to different phonetic inventories) or task-agnostic continual learning (where tasks are not predefined). This will allow us to investigate the effectiveness of our method in handling more complex scenarios and potentially lead to a more robust continual learning for ASR in machine speech chain framework.

\section{Ethics Statement}
\label{ethics-statement}
Our study followed the scientific methodology and ethics. The LJ Speech dataset that we used is a public domain dataset which is not in violation of license and data ethics. LJ Speech dataset is an English language speech dataset consisting of 13,100 short audio clips of a single speaker reading passages from 7 non-fiction books. The audio part was recorded and donated voluntarily by the speaker to the public domain. The texts that were read by the speaker are also in the public domain. We are aware of the usage of synthetic data that is generated by text-to-speech (TTS) to assist the continual learning of automatic speech recognition (ASR). There is potential to perpetuate ethical risk, such as bias and attribution issues in the synthetic samples. However, our proposed method utilizes TTS within a closed-loop framework, allowing us to better control the generation process and mitigate such issues. Furthermore, we believe this method can alleviate key challenges, such as the reliance on large quantities of real human speech data. 

\section{Acknowledgements}
\label{sec:acknowledgements}
Part of this work is supported by JSPS KAKENHI Grant Numbers JP21H05054 and JP23K21681, as well as JST Sakura Science Program.

\bibliographystyle{IEEEbib}
\bibliography{main}

\end{document}